\documentclass[conference]{IEEEtran}
\IEEEoverridecommandlockouts

\usepackage{cite}
\usepackage{amsmath,amssymb,amsfonts}
\usepackage{graphicx}
\usepackage{textcomp}
\usepackage{xcolor}
\usepackage{booktabs}
\usepackage{url}
\def\BibTeX{{\rm B\kern-.05em{\sc i\kern-.025em b}\kern-.08em
    T\kern-.1667em\lower.7ex\hbox{E}\kern-.125emX}}

\graphicspath{{figures/}}

\begin{document}

\title{LoMASA: A Local Multi-Agent Sales Assistant with Confidence-Gated Delegation for B2B Quoting}

\author{
  \IEEEauthorblockN{
    1\textsuperscript{st} Yu Liu\textsuperscript{\dag},~
    2\textsuperscript{nd} Xingbang Du\textsuperscript{\ddag}
  }
  \IEEEauthorblockA{
    \textsuperscript{\dag}\textit{Xi'an International Studies University}, Xi'an, China\\
    \textsuperscript{\ddag}\textit{Information Systems Design Laboratory,
    Hokkaido University}, Sapporo, Japan\\
    liuyu021027@gmail.com,~
    xingbang.du.d1@elms.hokudai.ac.jp
  }
}

\maketitle

\begin{abstract}
Frontline sales staff spend much of their day re-coordinating quotes after customers change quantities, ask
for discounts or new payment terms, or costs move. We present LoMASA (Local Multi-Agent Sales Assistant), a training-free multi-agent system
that runs entirely on one desktop computer. A 35B mixture-of-experts model (System-1) makes each decision in a single forward
pass and reads its confidence from the probabilities of the answer options. A 27B dense model (System-2)
writes customer replies and escalation briefs and is consulted when System-1 is unsure. Code computes every
price, margin and option, and hard pricing rules cannot be overridden. On a test split of 500
simulated episodes with 1512 decisions built from public supply-chain data, the salesperson has to
decide only 19.5\% of decisions: 38.0\% are settled automatically, with a median
latency of 2.67 seconds, and 42.5\% are escalated to a manager by the hard rules. On
the automatically settled decisions LoMASA is as accurate as a rule baseline (0.894 vs.\
0.894) and makes fewer commitments whose margin later falls below the floor (2.61\%
vs.\ 4.53\%); the price is additional escalations to the manager. Option probabilities are
well calibrated without any training, unlike confidence stated in words by the same model. We also report
where the design does not help: the 27B model rarely changes a decision.
\end{abstract}


\section{Introduction}
B2B quoting is a coordination task. Every customer message or cost change forces a salesperson to recompute
the quote, check it against pricing policy, decide whether to commit, and then write a reply or ask a manager.
Most of these decisions are routine; a few are risky. An assistant is useful if it takes over the routine
ones, reliably hands over the risky ones, and never makes the outcome worse. Field studies show that
generative AI can raise the productivity of customer-facing and knowledge workers
\cite{brynjolfsson2025generative,dell2023navigating}, but many sales teams cannot send customer and pricing
data to a cloud service, and small teams rarely have the labelled data needed to train a model.

We build such an assistant, LoMASA (Local Multi-Agent Sales Assistant), from two open models on one Mac mini.
LoMASA is a multi-agent system: an intake agent, a fast decision agent, a slower reviewing and writing agent,
a deterministic harness and the human salesperson each have a fixed role. We evaluate it against two
requirements that must hold together: (i) it saves the salesperson a substantial share of decisions, and (ii) it introduces no
negative effects on accuracy or risk compared with following the pricing rules. We also report the one cost we
found, extra escalations to the manager. Our contributions are:
\begin{enumerate}
\item \textbf{Confidence-gated delegation without training.} System-1 reads both the decision and its
  confidence from option log-probabilities in one forward pass; thresholds solved on validation for a target
  error rate decide whether the system acts or defers. These confidences are calibrated out of the box,
  unlike verbalised confidence from the same model.
\item \textbf{Code computes, the language model judges.} All numbers come from code, every number in outgoing
  text is checked against code-computed values, and hard rules cannot be bypassed. This is the structural
  reason the assistant cannot commit against the pricing policy.
\item \textbf{Case retrieval by numeric features.} Retrieving similar past cases by standardised economic
  features, rather than by text embeddings, substantially improves escalation decisions.
\item \textbf{A reproducible benchmark and an honest evaluation} on public data, including negative results:
  on this task the 27B model adds little to the decisions themselves.
\end{enumerate}

\section{Related Work}
\textbf{Confidence of language models.} Modern neural networks are often miscalibrated, and temperature
scaling is a simple post-hoc fix \cite{guo2017calibration}. Large language models can be reasonably
calibrated at the token level on multiple-choice questions \cite{kadavath2022language}. Confidence can also
be elicited in words \cite{lin2022teaching,tian2023just}, but verbalised confidence tends to be
overconfident \cite{xiong2024can}. We use option log-probabilities and compare them with verbalised
confidence from the same model.

\textbf{Selective prediction and deferral.} Selective classification abstains when confidence is low
\cite{geifman2017selective}, and learning-to-defer methods train a model to hand cases to a human expert
\cite{madras2018predict,mozannar2020consistent}. We apply a training-free version: thresholds on calibrated
confidence chosen for a target error rate.

\textbf{Fast and slow models.} The System-1/System-2 distinction \cite{kahneman2011fast} has inspired
agents that combine a fast and a slow model \cite{lin2023swiftsage} and methods that distil slow reasoning
into fast models \cite{yu2024distilling}. Model cascades and routers send easy queries to cheap models and
hard ones to expensive models \cite{chen2023frugalgpt,yue2024large,ong2025routellm}; reasoning in text can
improve accuracy on multi-step problems \cite{wei2022chain}. Our cascade adds a human as the final stage and
asks whether the slow model actually changes decisions.

\textbf{Agents, tools and retrieval.} Language-model agents interleave reasoning with tool use
\cite{yao2022react,schick2023toolformer}, delegate arithmetic to programs \cite{gao2023pal}, and retrieve
external knowledge \cite{lewis2020retrieval}. Multi-agent frameworks coordinate several such agents
\cite{wu2023autogen,hong2024metagpt,park2023generative}. We keep the agent structure deliberately simple and
let code own all numbers.

\section{Methodology}
This section first formalises the quoting task (Sections~\ref{sec:formulation}--\ref{sec:data}) and then
describes LoMASA, whose components map one-to-one onto the quantities of the task
(Sections~\ref{sec:arch}--\ref{sec:grounding}).

\begin{figure*}[t]
\centering
\includegraphics[width=\textwidth]{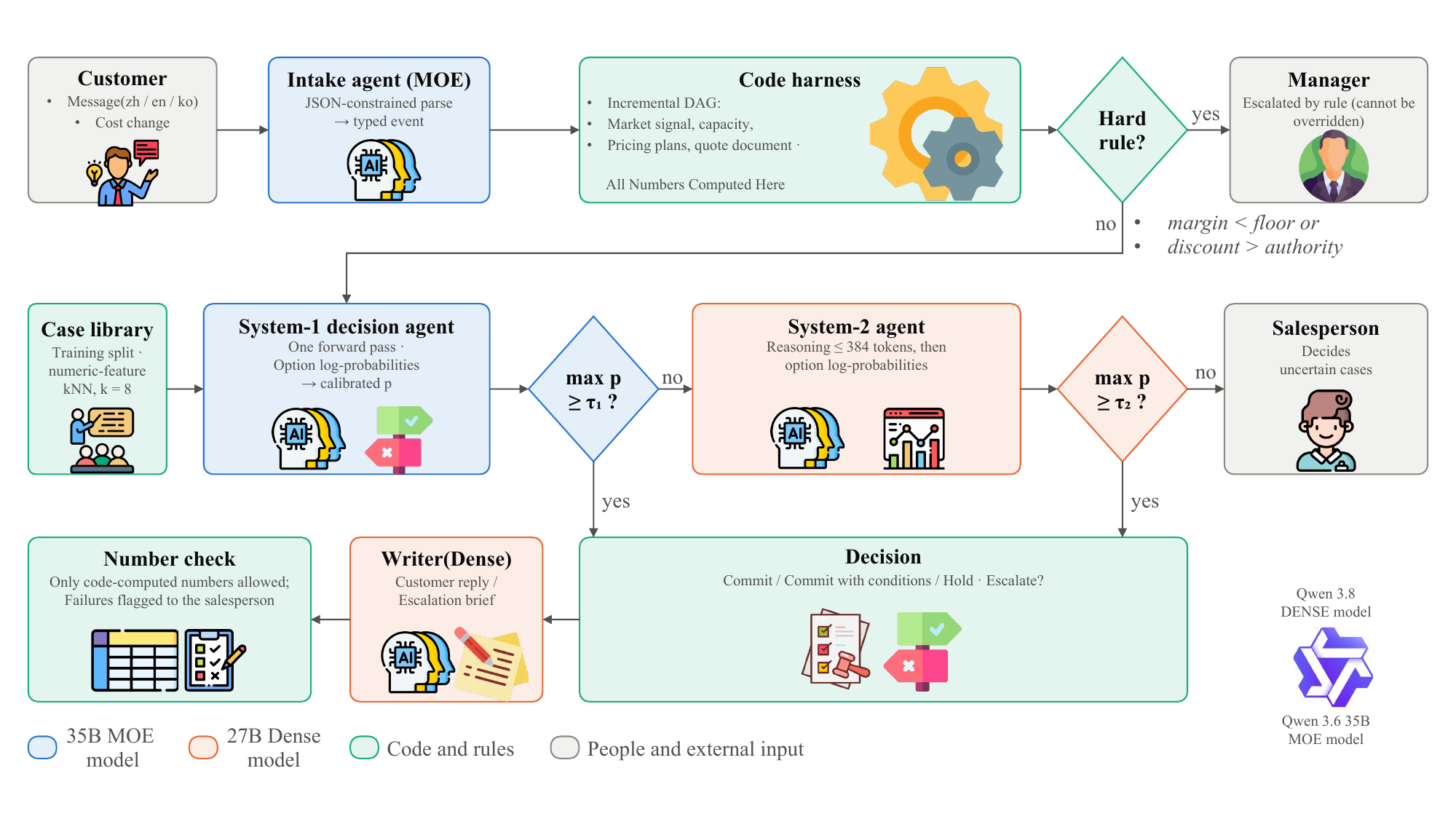}
\caption{LoMASA overview. Code computes all numbers and enforces hard rules; System-1 decides when confident;
System-2 is consulted otherwise; uncertain cases go to the salesperson. The 27B model also drafts outgoing
texts, which may contain only code-computed numbers. (Placeholder figure.)}
\label{fig:system}
\end{figure*}

LoMASA\subsection{Problem Formulation}\label{sec:formulation}
A quote is a state
\begin{equation}
q=(s,\,Q,\,P,\,d,\,t,\,D,\,c,\,\boldsymbol{\ell},\,\kappa),\label{eq:state}
\end{equation}
where $s$ is the product, $Q$ the quantity, $P$ the list unit price, $d$ the discount, $t$ the payment terms in
days, $D$ the delivery time in days, $c$ the current cost index, $\boldsymbol{\ell}=(\ell_1,\ell_2,\ell_4,\ell_8)$
the cost index $1$, $2$, $4$ and $8$ days ago, and $\kappa$ the daily capacity. An episode starts from an
initial quote $q_0$ and receives a sequence of events $e_1,\dots,e_n$ with $n\in\{2,3,4\}$. Each event has a
type in $\mathcal{E}=\{$quantity change, counter-offer, cost jump, payment change, delivery change$\}$ and a
payload, and it updates the state deterministically:
\begin{equation}
q_\tau = f(q_{\tau-1}, e_\tau),\qquad \tau=1,\dots,n.\label{eq:transition}
\end{equation}
After every event the assistant must output a decision pair
\begin{equation}
\begin{aligned}
y_\tau &= (a_\tau,\,b_\tau),\qquad b_\tau\in\{0,1\},\\
a_\tau &\in \mathcal{A}=\{\textsc{commit},\,\textsc{cond},\,\textsc{hold}\},
\end{aligned}\label{eq:decision}
\end{equation}
where $a_\tau$ states whether to commit, commit with conditions, or hold, and $b_\tau=1$ means escalate to a
manager. The decision has to be made with the information available at time $\tau$; the correct answer
depends on tomorrow's cost $c^{+}$, which is unknown.

\subsection{Quote Economics and Pricing Policy}\label{sec:economics}
A policy $\pi=(m^{\star},\,m^{\min},\,\bar d,\,r)$ fixes a target margin $m^{\star}$, a floor margin $m^{\min}$,
the sales authority $\bar d$ (largest discount a salesperson may grant) and an annual financing rate $r$; we
use $m^{\star}=0.15$, $m^{\min}=0.05$, $\bar d=0.08$ and $r=0.06$. With a product-specific factor $k_s$, the
unit cost is $u=k_s\,c$. The effective price, financing cost and margin are
\begin{align}
p &= P\,(1-d), \qquad \phi = p\,r\,\frac{t}{365},\label{eq:price}\\
m(q;c) &= \frac{p-u-\phi}{p}.\label{eq:margin}
\end{align}
Solving $m=m^{\min}$ for $p$ gives the floor price
\begin{equation}
p^{\min}=\frac{u}{1-m^{\min}-r\,t/365}.\label{eq:floor}
\end{equation}
Capacity is sufficient if $Q\le \kappa D$.

\subsection{Hard Rules and Labels}\label{sec:labels}
Two situations require a manager regardless of any model output:
\begin{equation}
H(q)=\mathbf{1}\big[\,m(q;c)<m^{\min}\;\lor\; d>\bar d\,\big].\label{eq:hard}
\end{equation}
The commit label is the policy decision evaluated at the realised next-day cost $c^{+}$:
\begin{equation}
a^{\ast}=\begin{cases}
\textsc{commit} & \text{if } m(q;c^{+})\ge m^{\star} \text{ and } Q\le\kappa D,\\
\textsc{cond} & \text{else if } m(q;c^{+})\ge m^{\min},\\
\textsc{hold} & \text{otherwise,}
\end{cases}\label{eq:commitlabel}
\end{equation}
and the escalation label is
\begin{equation}
b^{\ast}=H(q)\;\lor\;\mathbf{1}\big[m(q;c^{+})<m^{\min}\big].\label{eq:esclabel}
\end{equation}
The RULE baseline applies \eqref{eq:commitlabel} with the current cost $c$ instead of $c^{+}$ and escalates
exactly when $H(q)=1$. It is therefore always consistent with the policy, but it cannot anticipate cost
movements and has no notion of uncertainty. Decisions with $H(q)=1$ are called \emph{rule cases}; all other
decisions are \emph{non-rule cases}.

\subsection{Data Construction}\label{sec:data}
\textbf{Costs.} The cost index path comes from scm1d \cite{spyromitros2016multi} on OpenML
\cite{vanschoren2014openml}, derived from the TAC supply-chain management game \cite{collins2004supply}. For
eight products it provides the current price, the lags $\boldsymbol{\ell}$ and the next-day price, which we use
as $c$, $\boldsymbol{\ell}$ and $c^{+}$. A cost-jump event can only occur on days where the price moved by more
than $5\%$; such days are rare, so cost jumps account for only 12 of the 1512 test
decisions (counter-offers 541, quantity changes 429, payment changes
284, delivery changes 246). All events of an episode fall on the same
scm1d day, so one next-day cost $c^{+}$ serves every decision in the episode.

\textbf{Customers and orders.} Customer tiers and order quantities come from Online Retail II
\cite{online_retail_ii_502}; scheduled shipping days, payment types and requested discount rates come from
DataCo \cite{constante2019dataco}. The initial unit price is the cost marked up by a factor drawn uniformly
from $[0.06,0.30]$, and the initial capacity covers $60\%$ to $180\%$ of the ordered quantity.

\textbf{Splits.} We generate 1000 training, 300 validation and 500 test episodes. Splits are disjoint by scm1d
game and by customer. Labels are computed by \eqref{eq:commitlabel}--\eqref{eq:esclabel}; test labels were
locked behind an access guard until the configuration was frozen.

\textbf{Messages.} Each event is delivered as a message in Chinese, English or Korean (from the customer, or
from a colleague for cost changes), rendered from templates that the 27B model paraphrased offline, and
perturbed with varied number formats, percent wording, greetings, casing and punctuation.

\subsection{LoMASA Architecture}\label{sec:arch}
Fig.~\ref{fig:system} shows the pipeline. LoMASA consists of agents with fixed roles and a deterministic
harness that owns all numbers. Both language models run locally with llama.cpp \cite{llamacpp}:
\begin{itemize}
\item \textbf{Intake agent} (35B): parses a customer message into a typed event $e_\tau$ under a JSON schema.
\item \textbf{Harness} (code): applies \eqref{eq:transition}, computes \eqref{eq:price}--\eqref{eq:floor},
  checks capacity, enumerates feasible pricing plans and evaluates $H(q)$ in \eqref{eq:hard}.
\item \textbf{System-1 decision agent} (Qwen3.6-35B-A3B \cite{yang2025qwen3}, a 35B mixture-of-experts model
  with about 3B active parameters, 4-bit quantised): estimates $a_\tau$ and $b_\tau$ with confidences.
\item \textbf{System-2 agent} (Qwen3.8-27B, dense, 4-bit quantised): re-examines decisions that System-1 is
  unsure about, and drafts customer replies and escalation briefs.
\item \textbf{Validator} (code): checks every number in outgoing text.
\item \textbf{Salesperson} and \textbf{manager} (human): take the decisions that the models or the rules hand over.
\end{itemize}
The decision space of the agents is exactly $\mathcal{A}\times\{0,1\}$ from \eqref{eq:decision}, and the
harness evaluates the same functions $m$ and $H$ that define the labels, so the models never see a number that
the harness has not computed. Each decision therefore ends in one of three places: the manager (hard rule),
no one (settled automatically), or the salesperson (deferred).

\subsection{Incremental Recomputation}\label{sec:dag}
The harness is a dependency graph over five nodes: market signal, available-to-promise (capacity), pricing
plans, quote document and decision. Each event type invalidates only its descendants: a quantity change and a
delivery change affect capacity, quote document and decision; a counter-offer and a payment change affect
pricing, quote document and decision; a cost jump affects market signal, pricing and decision. At every step
the incrementally updated nodes are compared with a full recomputation of all nodes on the current state, so a
dependency error cannot pass unnoticed.

The pricing node enumerates plans $(d',t')$ with $d'\in\{0,0.005,\dots,\bar d\}$ and $t'$ in the allowed terms
$\{0,30,60,90\}$, keeps those with $m\ge m^{\min}$, and ranks them by closeness to the customer's request,
$10\,|d'-d|+|t'-t|/90$, then by margin. The top three plans are given to System-2 and listed in escalation briefs.

\subsection{Case Retrieval}\label{sec:rag}
Past decisions from the training split form a case library. Each decision point is described by a numeric
feature vector
\begin{equation}
\mathbf{x}(q,e)=\big(m,\,d,\,t,\,\min(\kappa D/Q,3),\,\delta_1,\delta_2,\delta_4,\delta_8,\,\beta,\,\mathbf{o}_e\big),\label{eq:features}
\end{equation}
where $\delta_h=(c-\ell_h)/\ell_h$ are cost changes, $\beta$ is the position of $c$ within the range of the last
eight days, and $\mathbf{o}_e$ is a one-hot encoding of the event type. The numeric features are standardised
with the training mean and standard deviation (the one-hot part is kept as is), giving $\tilde{\mathbf{x}}$,
and compared with a weighted distance
\begin{equation}
\mathrm{dist}(\mathbf{x},\mathbf{x}')=\big\|\mathbf{w}\odot(\tilde{\mathbf{x}}-\tilde{\mathbf{x}}')\big\|_2,\label{eq:dist}
\end{equation}
with weight $3$ on the margin, $1$ on discount, capacity and one-day change, $0.5$ on the other numeric
features and $0.3$ on the event type. The $k=8$ nearest cases are inserted into the prompt together with their
outcomes. We first tried text embeddings, but they matched surface wording (event type, customer tier) rather
than economics.

\subsection{Confidence from Option Log-Probabilities}\label{sec:conf}
The prompt for System-1 contains the policy $\pi$, a description of $q_\tau$ and $e_\tau$ written by code
(prices, cost, cost changes, margin from \eqref{eq:price}--\eqref{eq:margin} and its gaps to $m^{\star}$ and
$m^{\min}$, capacity), the retrieved cases, and a multiple-choice question whose options are single tokens. One forward pass yields the next-token
log-probabilities $z_i=\log P(\text{label}_i\mid\text{prompt})$ for the option labels, from which
\begin{equation}
p_i=\frac{\exp(z_i/T)}{\sum_{j}\exp(z_j/T)}.\label{eq:p}
\end{equation}
The commit question has three options ($\mathcal{A}$) and the escalation question two (yes or no), giving
$\mathbf{p}^{a}$ and $p^{b}=P(b=1)$. The temperature is fitted separately for each question on the validation
split by minimising the negative log-likelihood,
\begin{equation}
T=\arg\min_{T>0}\;-\frac{1}{N}\sum_{n=1}^{N}\log p_{y_n}(T).\label{eq:temp}
\end{equation}
No model weights are changed. Confidences are $\gamma^{a}=\max_i p^{a}_i$ and $\gamma^{b}=\max(p^{b},1-p^{b})$.

\subsection{Confidence-Gated Delegation}\label{sec:cascade}
Each decision passes through at most three deciders. With System-1 confidence $\gamma_1$ and System-2
confidence $\gamma_2$,
\begin{equation}
\text{decider}=\begin{cases}
\text{System-1} & \text{if } \gamma_1\ge\tau_1,\\
\text{System-2} & \text{else if } \gamma_2\ge\tau_2,\\
\text{salesperson} & \text{otherwise,}
\end{cases}\label{eq:route}
\end{equation}
applied separately to the commit and the escalation question. If either question is below $\tau_1$, System-2
is called once; it receives the same policy, state and cases plus the outputs of the capacity and pricing tools, reasons for at
most 384 tokens, then answers both multiple-choice questions and is scored with \eqref{eq:p}. Only the
question that was uncertain for System-1 takes System-2's answer. The thresholds are set on the
validation split for a target error rate $\epsilon=0.1$, as the smallest threshold whose accepted decisions
have an error rate of at most $\epsilon$:
\begin{equation}
\tau=\min\Big\{\theta:\;\frac{\sum_n \mathbf{1}[\gamma_n\ge\theta,\;\hat y_n\ne y_n]}{\sum_n\mathbf{1}[\gamma_n\ge\theta]}\le\epsilon\Big\}.\label{eq:tau}
\end{equation}
This gives $\tau_1=0.739$ (commit) and $0.619$ (escalation) for System-1 and $\tau_2=0.997$ for System-2.
Hard rules take precedence: if $H(q)=1$, the decision is escalated to the manager whatever the models output.
The final escalation decision is $b=H(q)\lor\mathbf{1}[p^{b}\ge 0.5]$. In our runs the models still scored
rule cases, so that the confidence estimates can be evaluated on all decisions; in deployment these outputs
are discarded.

\subsection{Grounded Text Generation}\label{sec:grounding}
After a decision, the 27B model drafts an escalation brief if the decision is escalated, or a customer reply
if it commits (with or without conditions), from the code-computed state and plans; a hold without escalation
needs no text. Let $\mathcal{N}(x)$ be the set of numbers in a text $x$ and $\mathcal{V}(q)$ the set of values the
harness computed for this decision (quantity, prices, discount, terms, delivery, margins, policy limits and
plan values, each also rounded to one decimal place and to an integer). A text passes the check only if
\begin{equation}
\forall v\in\mathcal{N}(x):\;\min_{u\in\mathcal{V}(q)\cup\{0,\dots,10\}}|v-u|\le 0.011,\label{eq:grounding}
\end{equation}
that is, every number in the text was computed by code (small counts are allowed); otherwise it is flagged
for the salesperson instead of being sent. In addition, System-1
scores each customer reply for over-commitment, i.e.\ promising more than the decision allows.

\section{Experimental Setup}
We compare five systems: RULE (pricing rules only), VERBAL (the 35B model states a confidence from 0 to 100),
S1 only (System-1 alone, without delegation), LoMASA (the full system), and ALL-27B (every decision by the 27B model). ALL-27B is run on the first 30
test episodes only, because each of its decisions takes about a minute. Metrics are accuracy, Macro-F1,
expected calibration error (ECE, 15 bins), area under the risk--coverage curve (AURC) and latency, with 95\%
cluster-bootstrap confidence intervals over episodes. All runs use one Mac mini M4 Pro with 64\,GB memory,
with greedy decoding and a single run per system. Decisions are evaluated on the true events, so that decision
errors are not mixed with parsing errors; the intake agent is evaluated separately on 600 messages.
Outgoing texts were generated for the last decision of each of the first 100 test episodes.

\section{Results}
\subsection{Time Saved}
Of the 1512 test decisions, 643 (42.5\%) trigger a hard rule and go to
the manager, 574 (38.0\%) are settled automatically, with neither question deferred,
and 295 (19.5\%) go to the salesperson. System-1 takes a median of
2.67\,s per automatic decision. Outgoing texts are drafted automatically: of 100 customer
replies and escalation briefs, 0 contained a number not computed by code and
0 were flagged as over-committing. We did not measure human handling time. If a decision
takes $T$ minutes by hand, the automatically settled decisions alone save $0.38\,T$ minutes per
decision, about 6.3 hours per 100 decisions at $T=10$. The salesperson is also spared the
rule cases, because code detects them and the 27B model drafts the brief; counting them as well gives an upper
bound of 13.4 hours, although the manager still has to decide those cases.

\subsection{No Negative Effects}
Table~\ref{tab:noharm} compares LoMASA with RULE on the same automatically settled decisions. Commit accuracy is
the same (difference 95\% CI [\textminus{}0.005, +0.005], so any loss is at most half a point), and
LoMASA makes significantly fewer commitments whose next-day margin falls below the floor (15 vs.\
26). The reduction comes from escalation: LoMASA escalated 53 of these
decisions to the manager, 11 of them justified. This is the cost of the design: over all
869 non-rule decisions, 90 escalations were unnecessary, which adds work for
the manager but no financial risk. Of 49 risky cases that no hard rule covers, LoMASA escalated
11 and deferred 23 to the salesperson, and missed 15; RULE
flagged none. By construction, a hard rule cannot be overridden and no automatic commitment can have a
current margin below the floor; the logs confirm this (0 and 0
violations). Incremental and full recomputation agreed on 100.0\% of steps.

\begin{table}[t]
\caption{Automatically settled decisions (574): LoMASA vs.\ RULE on the same cases}
\label{tab:noharm}
\centering
\begin{tabular}{lccc}
\toprule
 & LoMASA & RULE & Diff.\ 95\% CI\\
\midrule
Commit accuracy & 0.894 & 0.894 & [\textminus{}0.005, +0.005]\\
Below floor later & 2.61\% & 4.53\% & [\textminus{}3.08, \textminus{}0.91] pt\\
Escalated (justified) & 53 (11) & 0 (0) & --\\
\bottomrule
\end{tabular}
\end{table}

\subsection{Overall Comparison}
Table~\ref{tab:main} reports all 1512 test decisions, including those that the salesperson or the
manager takes over. Over all decisions LoMASA is close to RULE in commit quality (non-rule Macro-F1 difference
\textminus{}0.007, CI [\textminus{}0.015, +0.001]): the label depends on
tomorrow's cost, which neither system can know. The difference is that LoMASA knows when it is unsure (ECE
0.075 vs.\ 0.189), which is what makes safe delegation possible. RULE has higher
escalation accuracy because it never escalates a non-rule case: it has no false alarms but misses all
49 risky non-rule cases, ranking them at chance (AUROC 0.500). LoMASA ranks them
with AUROC 0.742 and catches part of them at the cost of the unnecessary escalations above.

\begin{table*}[t]
\caption{Test results on all 1512 decisions. ``Non-rule'' excludes decisions settled by a hard rule.}
\label{tab:main}
\centering
\begin{tabular}{lcccccc}
\toprule
System & Commit acc. & Commit Macro-F1 & Commit ECE & Escalate acc. & Non-rule escalate AUROC & Non-rule commit Macro-F1\\
\midrule
RULE    & 0.811 & 0.800 & 0.189 & 0.968 & 0.500 & 0.558\\
VERBAL  & 0.738 & 0.665 & 0.125 & 0.942 & 0.728 & 0.422\\
S1 only & 0.795 & 0.787 & 0.018 & 0.897 & 0.782 & 0.554\\
LoMASA    & 0.812 & 0.798 & 0.075 & 0.923 & 0.742 & 0.551\\
\bottomrule
\end{tabular}
\end{table*}

\begin{figure*}[t]
\centering
\includegraphics[width=0.42\textwidth]{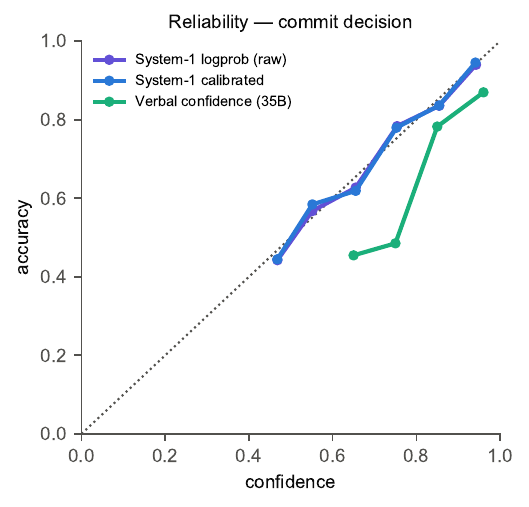}\hspace{0.04\textwidth}%
\includegraphics[width=0.42\textwidth]{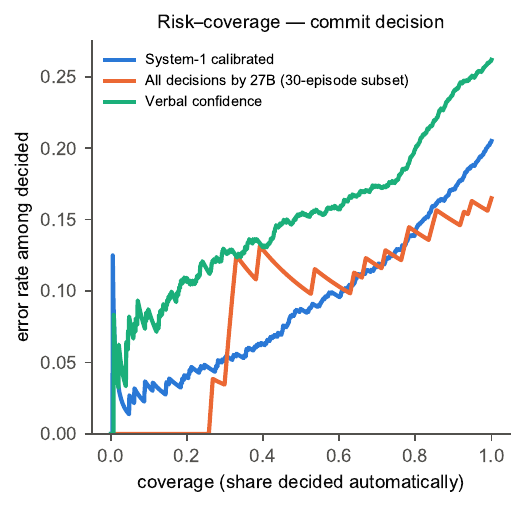}
\caption{Left: reliability of the commit decision. Right: risk--coverage of the commit decision (the 27B curve
uses the 30-episode subset).}
\label{fig:calib}
\end{figure*}

\subsection{Ablations}
\textbf{Confidence source.} Option log-probabilities are calibrated without training (commit ECE
0.016 raw, 0.018 after temperature scaling), whereas verbalised
confidence from the same model is not (ECE 0.125) and ranks errors worse (AURC
0.152 vs.\ 0.092); see Fig.~\ref{fig:calib}. Temperature scaling helps only for escalation
(ECE 0.032 to 0.016).

\textbf{Case retrieval.} Retrieving $k=8$ cases instead of none raises escalation accuracy from
0.716 to 0.897 and non-rule escalation AUROC from 0.631 to
0.782; commit accuracy is unchanged (0.800 vs.\ 0.795).

\textbf{System-2.} On the 30-episode subset (Table~\ref{tab:s2}) the cascade and ALL-27B make identical
decisions: on the 53 decisions that System-1 took itself, the 27B model chose the
same option in 53, and the rest were sent to the 27B model anyway. The cascade's
median latency is 2.74\,s against 46.93\,s. On the full test set,
System-2 decided only 1.5\% of commit decisions itself. The System-2 decision stage mostly adds
latency before a deferral and could be skipped in deployment; the value of the 27B model lies in the
human-facing writing.

\begin{table}[t]
\caption{ALL-27B subset (first 30 test episodes)}
\label{tab:s2}
\centering
\setlength{\tabcolsep}{4pt}
\begin{tabular}{lcccccc}
\toprule
System & Acc. & ECE & Esc.\ acc. & p50 (s) & p95 (s) & 27B calls/ep.\\
\midrule
LoMASA    & 0.835 & 0.079 & 0.907 & 2.74 & 50.62 & 1.50\\
ALL-27B & 0.835 & 0.102 & 1.000 & 46.93 & 49.45 & 3.23\\
\bottomrule
\end{tabular}
\end{table}

\textbf{Incremental recomputation.} The dependency graph avoids 40.0\% of node recomputations, but
wall-clock time per event is unchanged (2.27\,s vs.\ 2.25\,s) because rewriting the quote
note with the 27B model dominates and almost every event touches the quote.

\textbf{Intake.} Parsing reached 1.000 exact match on 600 messages in Chinese, English
and Korean. The messages come from templates with one field each, so this is an upper bound.

\subsection{Multilingual Input and Token Cost}
Tokenizers trained mostly on English can need several times more tokens for the same content in other
languages, which raises cost and latency for non-English users \cite{petrov2023language,ahia2023all}.
Table~\ref{tab:lang} breaks down token counts and latency by the language of the customer message.

\textbf{Input side.} With a vocabulary of 248{,}320 tokens, the tokenizer encodes the messages compactly
in all three languages: a message takes on average 22.5 tokens in Chinese, 23.7 in
English and 25.8 in Korean. The per-language messages are not parallel translations, but they carry
the same single field, so these counts give a rough comparison. Intake latency is practically the same across
languages (1194, 1195 and 1204\,ms),
because the message is short compared with the instructions and the JSON schema.

\textbf{Decision side.} Decisions show no language effect at all (median 2425,
2423 and 2425\,ms for System-1; 29.40,
29.28 and 29.39\,s for System-2 reasoning). This follows from the design:
the intake agent turns each message into a typed event, the harness writes the decision prompt in one working
language, and System-2 reasons in that language. The customer's language therefore never enters the decision
prompt, and the token tax cannot affect decision speed or quality.

\textbf{Customer replies.} The token tax appears where the model must generate text in the customer's
language. Korean replies were 31\% longer in tokens than Chinese replies (English:
13\%), and their median generation time rose from 4.71\,s to
5.95\,s. All replies were written in the customer's language. With only
5, 7 and 17 replies per language, this is an indication rather
than a measured effect. Because the tax is confined to the last, short generation step, its cost is limited to this
difference in reply time and does not change the time saved by automated decisions.

\textbf{A linguistic reading.} This pattern is consistent with the typology of the languages: Chinese is an
isolating language whose characters each carry a morpheme, whereas Korean is agglutinative and marks grammatical
relations and politeness with particles and verb endings attached to the stem \cite{sohn2001korean}, which
yields more word forms and hence more subword pieces \cite{sennrich2016neural}. Formal business replies use
the deferential speech level with longer honorific endings, which may explain why the overhead is larger for
replies than for the short incoming requests.

\begin{table}[t]
\caption{Token count and latency by customer language}
\label{tab:lang}
\centering
\setlength{\tabcolsep}{3.5pt}
\begin{tabular}{lccc}
\toprule
 & Chinese & English & Korean\\
\midrule
Message tokens (mean) & 22.5 & 23.7 & 25.8\\
Characters per token & 1.44 & 3.73 & 1.82\\
Intake latency, p50 (ms) & 1194 & 1195 & 1204\\
System-1 decision, p50 (ms) & 2425 & 2423 & 2425\\
System-2 reasoning, p50 (s) & 29.40 & 29.28 & 29.39\\
\midrule
Customer replies (n) & 5 & 7 & 17\\
Reply output tokens (mean) & 76 & 85 & 99\\
Reply generation, p50 (s) & 4.71 & 4.70 & 5.95\\
\bottomrule
\end{tabular}
\end{table}

\section{Limitations}
Episodes are simulated from public data rather than taken from real sales logs. The non-inferiority margin
was not fixed in advance; we report the confidence interval instead. LoMASA trades missed risks for
unnecessary escalations, and we did not model the manager's cost of handling them. Tomorrow's cost is partly
unpredictable, which caps attainable accuracy. Human handling time was not measured, so time savings are
given per assumed minute. We used one hardware setup and a single deterministic run with bootstrap intervals
instead of repeated runs. ALL-27B covers only 30 episodes, and we did not test whether the 35B model could
write the outgoing texts equally well. Multilinguality enters only message parsing and reply writing: the
messages are templated rather than free-form and not parallel across languages, the reply sample per language
is small, and we did not compare working languages for the decision prompt and the reasoning.

\section{Conclusion}
LoMASA, a local, training-free multi-agent assistant built from two open models, leaves the salesperson
19.5\% of quote decisions: it settles 38.0\% automatically with second-level latency,
hands rule cases to the manager with a drafted brief, and does so without lower accuracy or riskier commitments
than following the pricing rules, at the cost of extra escalations. The enabling pieces are simple: confidences read from
option log-probabilities, code for every number, and hard rules that the model cannot override.

\section*{Data Availability}
Online Retail II \cite{online_retail_ii_502} and DataCo Smart Supply Chain \cite{constante2019dataco} are
released under CC BY 4.0; scm1d is available on OpenML (dataset 41485). TODO: code repository link.

\bibliographystyle{IEEEtran}
\bibliography{ref}

\end{document}